\documentclass{article}

\usepackage{arxiv}
\renewcommand{\date}[1]{} 
\usepackage[utf8]{inputenc}   
\usepackage[T1]{fontenc}      
\usepackage{CJKutf8}          

\usepackage[utf8]{inputenc} 
\usepackage[T1]{fontenc}    
\usepackage{hyperref}       
\usepackage{url}            
\usepackage{booktabs}       
\usepackage{amsfonts}       
\usepackage{nicefrac}       
\usepackage{microtype}      
\usepackage{lipsum}
\usepackage{graphicx}
\graphicspath{ {./images/} }
\usepackage{booktabs} 
\usepackage{subcaption}
\usepackage{appendix}

\title{A Preview of XIYAN-SQL: A Multi-Generator Ensemble Framework for Text-to-SQL}

\author{Yingqi Gao, Yifu Liu\thanks{co-first author}, Xiaoxia Li,  Xiaorong Shi, Yin Zhu, Yiming Wang, Shiqi Li, Wei Li, Yuntao Hong, \and \textbf{Zhiling Luo}\thanks{Corresponding Author, godot.lzl@alibaba-inc.com}, \textbf{Jinyang Gao}, \textbf{Liyu Mou}, \textbf{Yu Li}
\\
\\ Alibaba Group
\\ \url{https://github.com/XGenerationLab/XiYan-SQL}}

\begin{document}
\begin{CJK}{UTF8}{gbsn}
\maketitle

\begin{figure}[htp] 
    \centering
    \includegraphics[width=0.75\textwidth]{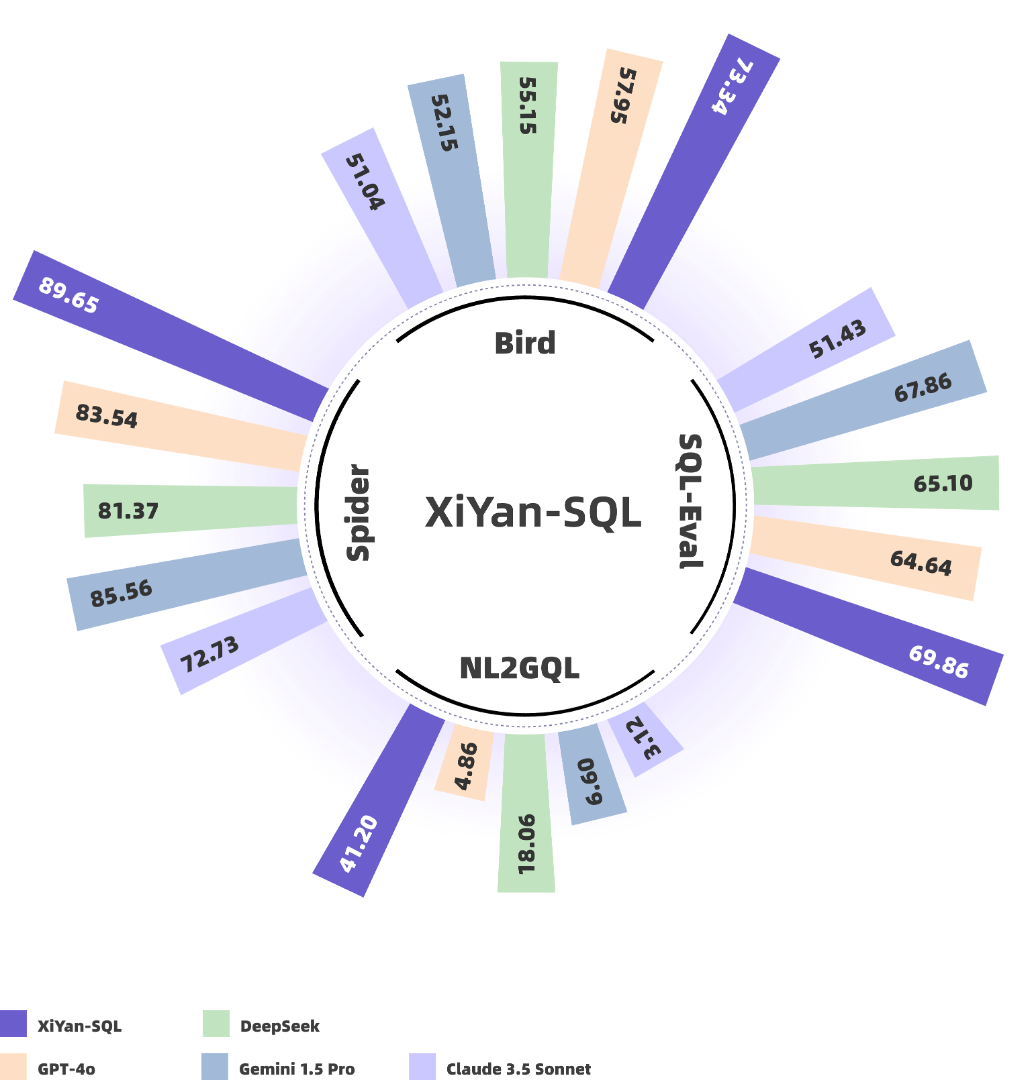}
    \label{fig:xiyansql}
\end{figure}

\newpage
\begin{abstract}
To tackle the challenges of large language model performance in natural language to SQL tasks, we introduce XiYan-SQL, an innovative framework that employs a multi-generator ensemble strategy to improve candidate generation.
We introduce M-Schema, a semi-structured schema representation method designed to enhance the understanding of database structures.
To enhance the quality and diversity of generated candidate SQL queries, XiYan-SQL integrates the significant potential of in-context learning (ICL) with the precise control of supervised fine-tuning.
On one hand, we propose a series of training strategies to fine-tune models to generate high-quality candidates with diverse preferences.
On the other hand, we implement the ICL approach with an example selection method based on named entity recognition to prevent overemphasis on entities.
The refiner optimizes each candidate by correcting logical or syntactical errors.
To address the challenge of identifying the best candidate, we fine-tune a selection model to distinguish nuances of candidate SQL queries.
The experimental results on multiple dialect datasets demonstrate the robustness of XiYan-SQL in addressing challenges across different scenarios.
Overall, our proposed XiYan-SQL achieves the state-of-the-art execution accuracy of 75.63\% on Bird benchmark, 89.65\% on the Spider test set, 69.86\% on SQL-Eval, 41.20\% on NL2GQL.
The proposed framework not only enhances the quality and diversity of SQL queries but also outperforms previous methods.


\end{abstract}


\section{Introduction}
The ability to convert natural language queries into structured query language (SQL) through natural language to SQL (NL2SQL) technology represents a significant advancement in making complex datasets more accessible.
It greatly facilitates both non-expert and advanced users in extracting valuable insights from extensive data repositories~\cite{chen2023text, liu2024survey, tai2023exploring, rat_sql, gu2023few, bird, liu2022semantic, wang2022proton, sqlgen, dts_sql, sun2023sql, rai2023improving}.
Recent advancements in large language models (LLMs) have significantly enhanced the efficacy and accuracy of NL2SQL applications.


There are generally two approaches for NL2SQL solutions based on LLMs: prompt engineering~\cite{dong2023c3, DAIL, chasesql, dinsql}, and supervised fine-tuning (SFT)~\cite{codes}. 
Prompt engineering leverages the intrinsic capabilities of the model by optimizing prompts to generate diverse SQL queries.
Prompt engineering has demonstrated promising results in NL2SQL using zero-shot~\cite{dong2023c3} or few-shot prompting~\cite{macsql, DAIL, dinsql}.
This type of approach typically employs closed-source models with enormous parameters, such as GPT-4~\cite{gpt4} and Gemini 1.5~\cite{gemini15}, which present significant potential and powerful generalization capability. 
However, most methods rely on multi-path generation and selecting the best option utilizing self-consistency, resulting in significant inference overheads. 
Approaches based on SFT seek to fine-tune models with much smaller parameter sizes on the NL2SQL task to produce more controllable SQL queries, such as CodeS~\cite{codes}. Nevertheless, due to their limited parameters, these methods struggle to perform complex NL2SQL reasoning and transfer to databases within a new domain.

In this technical report, we propose XiYan-SQL, a novel NL2SQL framework that employs a multi-generator ensemble strategy to enhance candidate generation. 
XiYan-SQL combines prompt engineering and the SFT method to 
generate candidate SQL queries with high quality and diversity.
To enhance high quality, we take advantage of the high controllability of SFT and utilize a range of training strategies to specifically fine-tune models to generate candidates with different preferences.
We introduce a two-stage multi-task training approach, which first activates the model's fundamental SQL generation capabilities, and subsequently transitions to a model with enhanced semantic understanding and diverse stylistic preferences.
To enhance diversity of generated candidates and capability of generating complex SQL queries, we utilize in-context learning to prompt LLMs. 
We propose to extract the skeleton of the questions by masking the named entities with common special tokens and using skeleton similarity to select and organize useful examples.
Then, each generator is followed by a refiner to correct logical or syntactical error based on execution results or error information.
Finally, a selection agent is required to select the best option.
Most existing works use self-consistency, but the most consistent result is not always the correct case.
So we propose to fine-tune a model to understand and identify the subtle differences among candidates and pick the final response.


Additionally, to enhance LLMs for better understanding of the database schema, we propose a new schema representation method named M-Schema.
Inspired by MAC-SQL Schema~\cite{macsql}, M-Schema presents the hierarchical structure between databases, tables, and columns in a semi-structured form.
We revised MAC-SQL Schema by adding data types and resulting in a more compact and clear format.
We conduct experiments to compare the impact of different schema representations on NL2SQL performance.
In comparison to DDL Schema and MAC-SQL Schema, LLMs using M-Schema demonstrate superior performance.

We present comprehensive evaluations on both relational and non-relational databases, specifically focusing on prominent systems such as SQLite, PostgreSQL, and nGQL.
XiYan-SQL demonstrates remarkable performance across a range of benchmarks, achieving the state-of-the-art performance on the Spider~\cite{spider}, SQL-Eval, and NL2GQL~\cite{r3_nl2gql}  datasets with 89.65\%, 69.86\%, and 41.20\% execution accuracy, respectively.
In the context of the more challenging Bird~\cite{bird} benchmark, XiYan-SQL also reaches the top accuracy of 75.63\%.
The impressive results achieved on various challenging NL2SQL benchmarks not only validate the effectiveness of our approach but also demonstrate its significant potential for broader applications in NL2SQL translation tasks.
XiYan-SQL can be accessed from 
\url{https://bailian.console.aliyun.com/xiyan}.
We also release the source code for connecting to the database and building M-Schema at~\url{https://github.com/XGenerationLab/M-Schema}.

\section{Overall Framework}
\begin{figure}[!tp]
    \centering
    \includegraphics[width=1.0\textwidth]{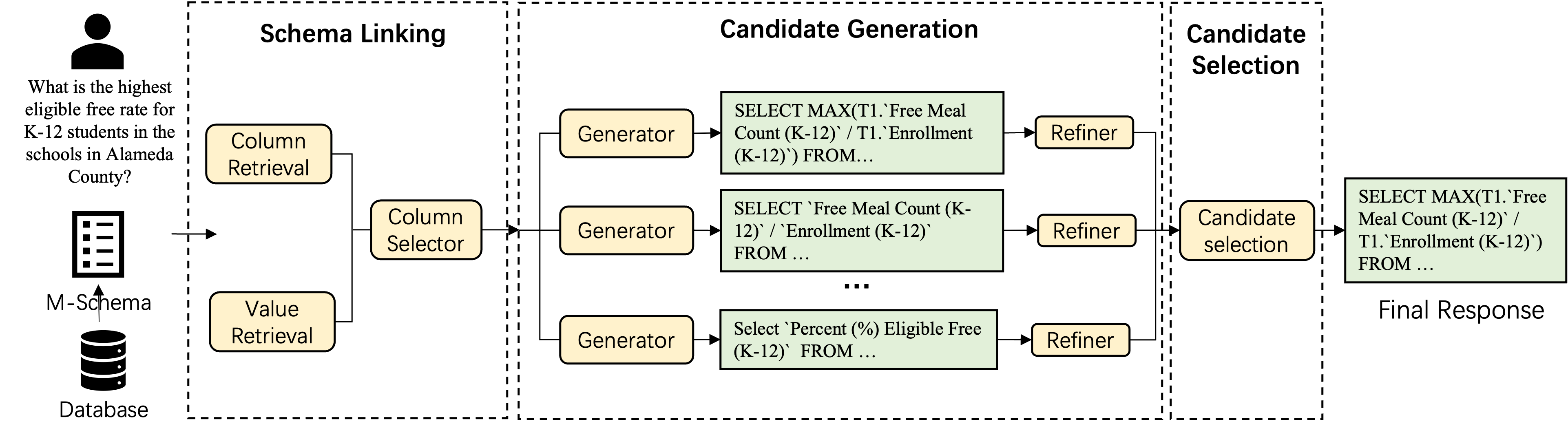}
\caption{Overview of the proposed XiYan-SQL workflow, which consists of three agents: 1) Schema Linking, which retrieves and selects the most database schema; 2) Candidate Generation: which generates high-quality candidate SQL queries using ICL and SFT generators; 3) Candidate Selection, which picks the final response among the generated candidates. M-Schema is served as schema representation and provided to LLMs.
\label{fig:pipeline}
}
\end{figure}

This section outlines the proposed XiYan-SQL framework, which consists of three primary components: 1) Schema Linking; 2) Candidate Generation; 3) Candidate Selection. 
Schema Linking is used to select relevant columns and retrieve values from a large database schema, helping to minimize irrelevant information and focus on related data. This contextual information is then organized into M-Schema and fed into Candidate Generation module to generate potential candidate SQL queries. These candidates are refined using a self-refinement process. Ultimately, a Candidate Selection agent compares all the candidates to determine the final SQL query .
This pipeline is illustrated in Figure~\ref{fig:pipeline}.

\section{M-Schema}
The database schema needs to be provided in the prompt so that LLM understands the database structure. We propose a novel representation named M-Schema.
M-Schema illustrates the hierarchical relationships between the database, tables, and columns in a semi-structured format and employs specifical tokens for identification: "【DB\_ID】" marks the database, "\# Table" signifies tables, and "【Foreign Keys】" indicates foreign keys.
For each table, we present table name and description, where table description can be omitted. 
The information from a table is converted into a list, where each item is a tuple representing the details of a column.
Each column includes the column name, data type, column description, primary key identifier, and example values.
Additionally, foreign keys need to be listed due to their importance.

\begin{figure}[!tp]
    \centering
    \includegraphics[width=0.9\textwidth]{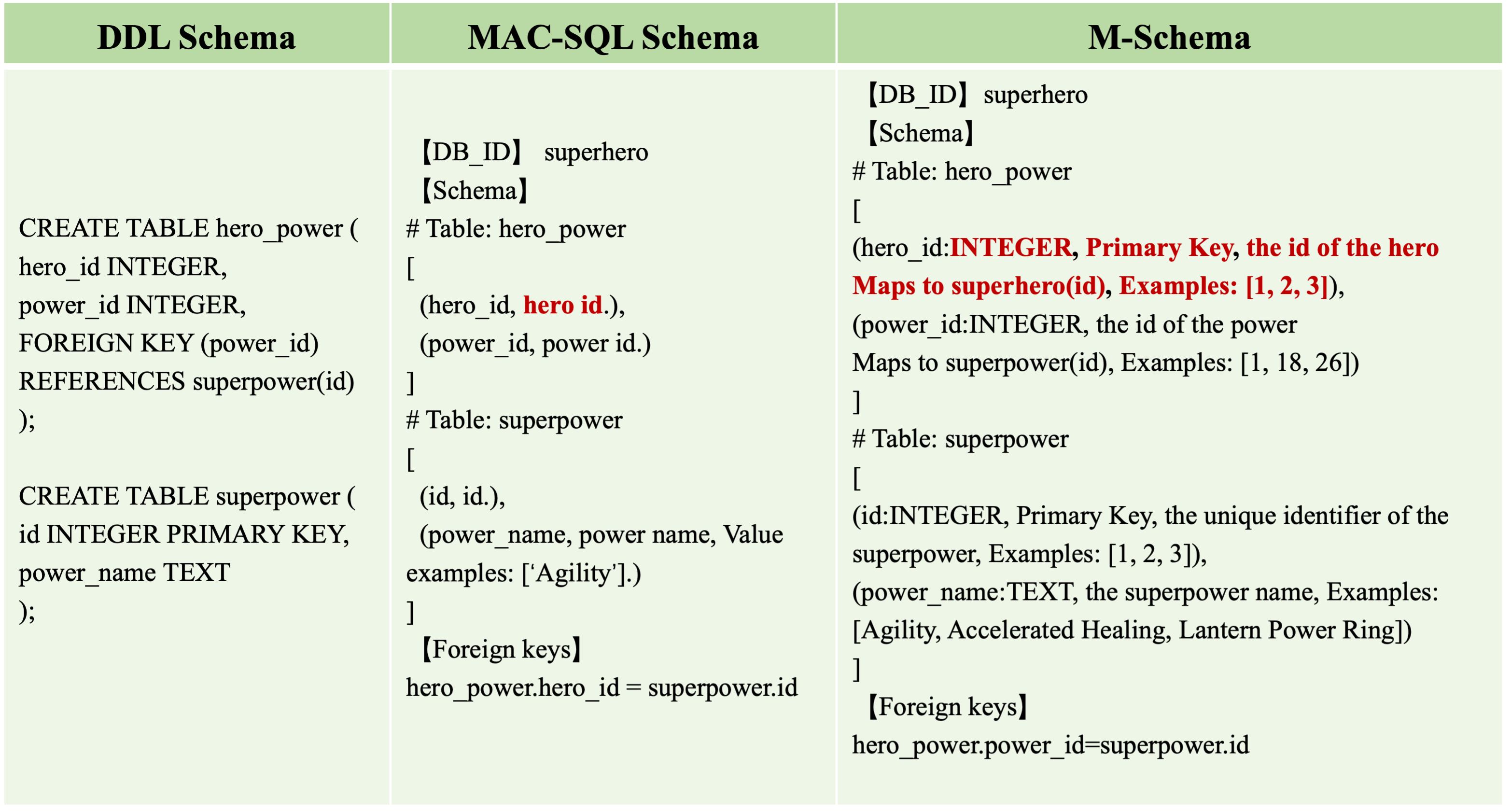}
    \caption{Examples of representing a database schema in DDL Schema, MAC-SQL Schema and M-Schema. The red text highlights the differences between M-Schema and MAC-SQL Schema. M-Schema adds data types, primary key markings, and changes the rules for displaying sample values.}
    \label{fig:schema_representation}
\end{figure}

Figure~\ref{fig:schema_representation} shows examples of representing a database in DDL Schema, MAC-SQL~\cite{macsql} Schema and M-Schema.
The Data Definition Language (DDL) schema is the most commonly used representation. However, it lacks essential table and column descriptions, as well as example values. Consequently, LLMs struggle to differentiate between similar columns.
Derived from MAC-SQL Schema, M-Schema is a more compact representation. It differs from MAC-SQL Schema mainly in column representation, detailed as follows:
\begin{itemize} 
    \item Data type. Data type ensures that the data is correctly structured and manipulated.
     MAC-SQL Schema lacks data type specifications, which may result in incorrect outcomes when generated SQL queries are executed.
    \item Primary key marking. We include primary key marking to maintain relationships between tables in a relational database.
    \item Column description. In MAC-SQL schema, the column description is derived from the column name, whereas M-Schema connects to the database to obtain more detailed descriptions.
    \item Value examples: We simplify "Value examples" marking into "Examples" to reduce redundancy. We also establish new display rules for values, such as string length and the number of examples.
\end{itemize}
Besides, the leading spaces in each column representation are removed from MAC-SQL Schema.
We release how to connect to the database engine and build the M-Schema representation at~\url{https://github.com/XGenerationLab/M-Schema} and support commonly used databases such as MySQL and PostgreSQL.

\section{Schema Linking}
Schema linking connects references in natural language queries to elements within a database schema, including table, columns and values. Our schema linking pipeline consists of a retrieval module and a column selector.

\textbf{Retrieval Module}
In order to search for similar values and columns in the database, similar to the approach in ~\cite{chasesql}, we first prompt the model with few-shot examples to identify keywords and entities in the question. 
We then use a column retriever to retrieve relevant columns. Based on the semantic similarity between the keywords and the column descriptions, we retrieve the top-k columns for each keyword.
To enhance efficiency, value retriever employs a two-phase retrieval strategy based on Locality Sensitive Hashing (LSH) and semantic similarity to identify similar values in the database.
The final selected schema is the union set of column retriever and value retriever.

\textbf{Column Selector}
Column Selector aims to reduce the tables and columns to minimally sufficient schema for SQL generation.
The retrieved schema from the previous step is organized as M-Schema and presented with LLMs.
We then employ a few-shot manner to prompt the language model to evaluating the relevance of each column to the user's query, selecting only those necessary.

\section{Candidate Generation}
For candidate generation, we employ various generators to generate high-quality and diverse SQL candidates. 
On one hand, we utilize a range of training strategies to specifically fine-tune the generation models, aiming to generate high-precision SQL candidates with diverse syntactic styles. 
On the other hand, we also incorporate the ICL approach to enhance the diversity of the SQL candidates. 
Our Refiner further improves the generated SQL queries. 
In the following sections, we provide a brief overview of each part.
\subsection{Fine-tuned SQL Generator}
\label{section51}
\begin{figure} 
    \centering
    \includegraphics[width=0.9\textwidth]{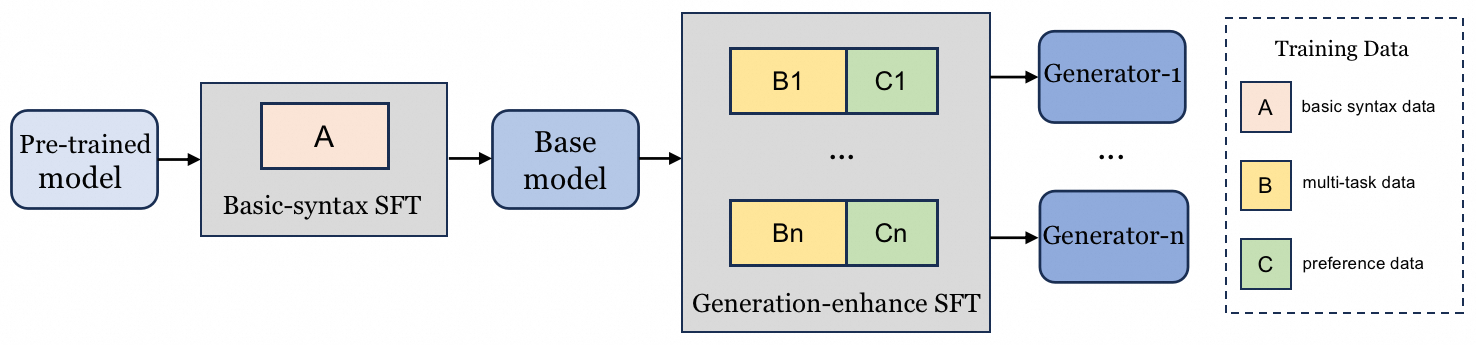}
    \caption{The two-stage and multi-task training pipeline for Fine-tuned SQL generators.}
    \label{fig3}
\end{figure}

The core purpose is to generate high-precision and diverse SQL candidates.
To this end, we take advantage of the high controllability of fine-tuning models on specific tasks to build a series of high-precision models with different preferences.
As shown in Figure \ref{fig3}, we employ a two-stage and multi-task training approach to fine-tune the model, including basic-syntax training and generation-enhance training.
Through this training approach, the intermediate and final results of our pipeline are a set of models with distinct advantages.

\textbf{Basic-syntax training}
Basic-syntax training focuses on fine-tuning the pre-trained model with the basic and single SQL patterns and syntax.
In this stage, the data used for training is SQL dialect-agnostic, covering basic syntax very comprehensively, with a total of tens of thousands of samples.
The training objective is to develop a base model that activates SQL generation capabilities and can serve as a transition to different specialized SQL tasks.

\textbf{Generation-enhance training}
After the first stage of training, we turn to generation-enhance training, aimed at enhancing the model's semantic understanding and stylistic preference in syntax.
In this stage, we can combine various multi-task data and syntactic preference data to obtain an enhanced model.
The model can benefit from multi-task data to better understand the mapping relationship between questions and SQL queries.
Specifically, in addition to the standard task of converting questions to SQL queries, we further design the task of converting SQL to questions, which aims to infer potential questions based on the provided contextual information and SQL query. 
We have defined the task from SQL to evidence, which is intended to select the most relevant evidence from a set of candidates based on the context and SQL.
Moreover, we also introduce the SQL discrimination and regeneration tasks, aimed at performing SQL optimization based on execution feedback, along with other related tasks.
This series of specialized tasks effectively enhances the linking between SQL and contextual information, thereby improving overall generation capabilities.
The model can benefit from various styles of patterns and syntactic features to better generate a wider diversity of SQL candidates.
We utilize different LLMs to rephrase the original query in multiple ways without altering its original meaning. 
This approach effectively expands the sample data into different syntactic styles, thereby teaching the model to learn from this data form during the training phase.

Due to multiple dialects in SQL queries, we can process each dialect separately during this stage, following this defined pipeline. Subsequently, we may opt to either train an individual model for each dialect or jointly train a multi-dialect model.
In practical applications, we can fine-tune a target model by selecting subsets of multi-task and preference data according to our needs, enabling the generation of high-quality SQL candidates.

\subsection{ICL SQL Generator}
\label{section52}
The performance of ICL-based NL2SQL generation depends not only on the inherent abilities of the model but also on the examples provided.
Several methods have been proposed to retrieve useful examples, such as masked question similarity and query similarity~\cite{DAIL}.
Although masked question similarity excludes the influence of table and column names, it is still sensitive to the entities.
Query similarity based method requires a preliminary model to generate an approximation SQL, so the capabilities of the preliminary model directly affect the final result.

XiYan-SQL employs an example selection strategy based on the skeleton similarity between the user question and the question from the training set.
All named entities in the question are first identified using NLTK’s tool, then the named entities of the same type are replaced with a special token.
For example, "China" and "America" are both identified as countries, so both of them are replaced by "<country>".
Other entities, such as enumeration values, are replaced by the column names.
This approach avoids focusing too much on entities, while the semantics of entities is preserved.
Then we compute embedding of modified questions in the training and test sets, and top-K examples from training sets that closely match the target question are selected.

Additionally, we noticed that SQL examples, which only manipulate one table, are of limited help for SQL generation involving multiple tables.
When selecting SQL examples, for questions that two or more tables are selected through schema linking, we only choose SQL queries that involve operations on multiple tables.
Based on the number of tables and the similarity threshold, a maximum of 5 examples are used for each question.

For benchmarks such as Bird and Spider, the databases of the training and test sets are not repeated, so presenting the schema of the examples in the prompt helps the model better understand the relationship between the schema and the SQL query.
In order to reduce token consumption and the interference of redundant columns, only the minimal set of columns is provided for each selected SQL example.


\subsection{SQL Refiner}
The generated candidate SQL queries inevitably contain logical or syntactical errors \cite{chasesql, chess}. 
By utilizing clues from these SQL query deficiencies, we can undertake corrections to some extent. 
To this end, we employ a SQL Refiner to optimize the generated SQL. 
In practice, based on schema-related context, the generated SQL queries, and execution results (including potential error information), we enable the model to perform a second round of corrective generation. 
The original SQL and the regenerated SQL can further be subjected to a selection model (as discussed in Section \ref{section6}) for optimal choice, and this process can be executed iteratively.

\section{Candidate Selection}
\label{section6}
Based on the schema linking and various candidate generators, we can generate a set of candidate queries for the given question. 
The challenge of selecting the correct and reasonable SQL query from the pool of candidates remains to be addressed.
Most methods \cite{mcssql,chess} employ self-consistency \cite{wang2023selfconsistency} to select the SQL query that appears most consistently across multiple candidate samples.
However, this approach has limitations: it cannot handle situations where none of the queries are consistent, and even the most consistent result is not always the correct case.

For this purpose, we employ a selection model to make judgments. 
We measure the consistency of SQL execution results to group them, allowing us to identify inconsistent samples from each group to form a candidate set.
Then, we utilize the selection model to select the most reasonable candidate based on the provided contextual information and the candidate set.
Instead of employing a prompt-based approach with LLM, we specifically fine-tune a model as a selection model to better distinguish nuances of candidate SQL queries. To align with the varying syntactic preferences of the SQL candidates, we also deliberately perform paraphrasing on the training data of the selection model.

\section{Experiments}
\subsection{Experimental Setup}
To assess the generalizability of the proposed XiYan-SQL framework, we evaluate it in an end-to-end way on both relational and non-relational graph databases.
Spider~\cite{spider} and Bird~\cite{bird} are widely-recognized cross-domain datasets that use SQLite. 
Since the test set of the BIRD benchmark is not available, we conduct experiments and performance evaluations on the development set.
SQL-Eval~\footnote{https://github.com/defog-ai/sql-eval} is an open-source PostgreSQL evaluation dataset released by Defog, constructed based on Spider.
NL2GQL~\cite{r3_nl2gql} built on graph databases is also involved in our experiments.
The detailed information of datasets is shown in Table~\ref{tab:dataset}.
We use Execution Accuracy (EX) to access the effectiveness of the generated SQL queries. 
EX compares the results of a predicted SQL query and a reference SQL query executed on a specific database instance.

\begin{table}
 \caption{Details of dataset used in our experiments.}
    \begin{center}
    \begin{minipage}{0.45\textwidth}
    \begin{tabular}{cccc}
    \toprule
    Dataset & Dialect  & \# Questions & \# DBs \\ \midrule
    Spider & SQLite & 1981 & 39  \\ 
    Bird & SQLite & 1534 & 11 \\ 
    SQL-Eval & PostgresQL & 304 & 11 \\ 
    NL2GQL& nGQL & 288 & 3 \\ \bottomrule
    \end{tabular}
    \end{minipage}
    \end{center}
    \label{tab:dataset}
\end{table}

\subsection{Bird Results}
We compare the performance of different NL2SQL methods on Bird benchmark in Table~\ref{tab:bird_result}.
XiYan-SQL reaches the top of Bird leaderboard with an accuracy of 75.63\%, outperforming the second place of 0.84\%.
CHASE-SQL~\cite{chasesql} framework employs multiple chain-of-thought prompting techniques to generate candidates, and subsequently implements a binary voting mechanism among 21 candidates, achieving an accuracy of 74.70\%. 
XiYan-SQL yields a competitive performance by voting among only 5 candidates.

We also observe that a significant number of the leading methods on the bird learderborad are based on prompt engineering techniques.
It suggests the immense potential of large-scale models and the importance of carefully designed prompts in optimizing model performance.
The SFT based method, ExSL + Granite-34B-Code, secures the fourth position with an accuracy of 73.17\%.
This notable performance indicates that, smaller-sized models are indeed capable of generating complex SQL queries effectively through advanced training techniques.
XiYan-SQL integrates the methodologies of SFT and ICL to balance the test time and the overall performance of the system.



\begin{table}
\centering
    \centering
    \begin{minipage}{0.53\linewidth}
    \caption{Performance comparison of different NL2SQL methods on Bird benchmark.}
    \vspace{0.5em}
    \label{tab:bird_result}
    \centering\begin{tabular}{ccc}
    \toprule
    Method & EX(Dev) & EX(Test)\\     
    \midrule
    CHASE-SQL + Gemini~\cite{chasesql} & 74.46 & 74.79 \\ 
    DSAIR + GPT-4o & 74.32 & 74.12 \\ 
    ExSL + granite-34b-code & 72.43 & 73.17  \\ 
    AskData + GPT-4o & 72.03 & 72.39\\ 
    OpenSearch-SQL, v2 + GPT-4o & 69.30 & 72.28\\ 
    Distillery + GPT-4o~\cite{distillery} & 67.21 & 71.83 \\ 
    CHESS~\cite{chess} & 68.31  & 71.10\\ 
    Insights AI & 72.16 & 70.26 \\ 
    PURPLE + RED + GPT-4o & 68.12 & 70.21 \\

    MCS-SQL~\cite{chess} & 63.36 & 65.45 \\ 
    SuperSQL~\cite{supersql} & 58.50 & 62.66 \\ 
    SFT CodeS-15B~\cite{codes} & 58.47 & 60.37 \\
    GPT-4o & 57.95 & - \\
    TA-SQL + GPT-4~\cite{tasql} & 56.19  & 59.14\\
    DAIL-SQL~\cite{DAIL} & 54.76 & 57.41\\
    \midrule
    XiYan-SQL & 73.34 & \textbf{75.63} \\ 
    \bottomrule
    \end{tabular}

    \end{minipage}
\hspace{0.01\linewidth} 
    \begin{minipage}{0.4\linewidth}
    \caption{Performance comparison of different NL2SQL methods on Spider test benchmark.}
    \vspace{0.5em}
    \label{tab:spider_result}
    \centering\begin{tabular}{cc}
    \hline
     \toprule
    Method & EX(\%) \\     
    \midrule
    MCS-SQL + GPT-4~\cite{mcssql} & 89.6 \\ 
    CHASE-SQL + Gemini 1.5~\cite{chasesql} & 87.6 \\ 
    PET-SQL~\cite{petsql} &  87.6 \\ 
    SuperSQL~\cite{supersql} & 87.0 \\
    DAIL-SQL + GPT-4~\cite{DAIL} & 86.6 \\ 
    DPG-SQL + GPT-4 & 85.6 \\ 
    Tool-SQL + GPT-4~\cite{tool_sql} & 85.6 \\
    DIN-SQL + GPT-4~\cite{dinsql} & 85.3 \\
    GPT-4o & 83.54\\ 
    C3 + ChatGPT + Zero-Shot~\cite{dong2023c3} & 82.3 \\ 
    \midrule
    XiYan-SQL & \textbf{89.65} \\ 
    \bottomrule
    \label{tab:results_spider_test}
    \end{tabular}

    \end{minipage}
\end{table}

\subsection{Spider Results}
To demonstrate the generalizability of our approach, we also evaluate XiYan-SQL on the Spider dataset.
As demonstrated in Table~\ref{tab:results_spider_test}, 
improvements in the underlying backbone model capabilities have contributed to notable improvements in performance metrics. 
Specifically, GPT-4o has achieved a remarkable accuracy of 83.54\%.
Moreover, XiYan-SQL refreshes the state-of-the-art execution accuracy of 89.65\%, with a marginal advantage of merely 0.05\% over previous leading models.

\begin{table}
 \centering

    
    \begin{minipage}{0.45\linewidth}
     \caption{Performance comparison of different methods on SQL-Eval benchamrrk.}
     \label{tab:result_sqleval}
     \vspace{0.5em}
    \centering\begin{tabular}{cc}
    \toprule
    Method & EX(\%)  \\    
    \midrule
    SQL-Coder-8B & 60.20 \\ 
    DeepSeek & 65.36 \\
    GPT-4o & 64.64 \\
    Claude 3.5 Sonnet & 51.43 \\
    Gemini 1.5 Pro & 67.86 \\
    \midrule
    XiYan-SQL & \textbf{69.86} \\    \bottomrule
    \end{tabular}
    \end{minipage}
\hfill
    \centering
    \begin{minipage}{0.45\linewidth}
    \caption{Performance comparison of different methods on NL2GQL benchmark.}
  \label{tab:result_graph_database}
   \vspace{0.5em}
    \centering\begin{tabular}{cc}
    \toprule
    Method & EX(\%)  \\   
    \midrule
    DeepSeek & 18.06 \\     
    GPT-4o & 4.86 \\
    Claude 3.5 Sonnet & 3.12 \\
    Gemini 1.5 Pro & 6.60 \\
    \midrule
    XiYan-SQL & \textbf{41.20} \\ 
    \bottomrule
    \end{tabular}
    \end{minipage}
\end{table}

\subsection{SQL-Eval Results}
Table~\ref{tab:result_sqleval} presents the results on SQL-Eval dataset. 
SQL-Eval provides multiple reference SQL queries and we choose the first option as groundtruth for metric computation.
XiYan-SQL reports the highest score of 69.86\% on SQL-Eval. 
We outperform SQL-Coder-8B~\footnote{https://huggingface.co/defog/llama-3-sqlcoder-8b} fine-tuned on LLaMA-3~\cite{llama3} by a large margin of 8.59\% and closed-source backbone models by 2$\sim$5 percent.
It demonstrates the generalizability of XiYan-SQL on SQL generation for PostgreSQL.

\subsection{NL2GQL Results}
To assess the effectiveness of XiYan-SQL on non-relational graph datasets, we sample a total of 288 examples from the NL2GQL~\cite{r3_nl2gql} dataset, which were previously utilized in MoMQ~\cite{2024arXiv241018406L}.
As shown in Table~\ref{tab:result_graph_database},
GPT-4o, DeepSeek, Gemini 1.5 Pro and Claude 3.5 Sonnet show a limited overall execution accuracy on NL2GQL dataset.
XiYan-SQL achieves 41.20\% execution accuracy, 
outperforming them by a large margin and demonstrating the best performance overall.

\subsection{Ablation Studies}
To further investigate the effectiveness of each component in our
framework,
we conduct ablation studies on the Bird development benchmark because of its challenging nature and more reflective of the real-world scenarios.

\subsubsection{M-Schema}
We conduct ablation study on the Bird development benchmark to present the impact of different schema representations on end-to-end SQL generation performance. 
To demonstrate the generalization ability of our proposed M-Schema, we use four powerful LLMs as NL2SQL generators, DeepSeek~\cite{liu2024deepseek},
Claude 3.5 Sonnet~\footnote{https://www.anthropic.com/news/claude-3-5-sonnet}, Gemini 1.5 Pro and GPT-4o~\cite{gpt4}.
As shown in Table~\ref{tab:ablation_schema}, all four models have performance improvements using M-Schema as the representation of database schema compared to DDL Schema, with an average increase of 2.03\%.
Although M-schema is similar to MAC-SQL Schema in structure, GPT-4o and Claude 3.5 Sonnet show 0.65\% and 0.78\% improvements, respectively.
While DeepSeek and Gemini 1.5 have slight accuracy decreases of 0.13\% and 0.26\%.
The experimental results indicate that M-Schema is a better representation than DDL Schema and MAC-SQL Schema and demonstrates powerful generalizability.

\begin{table}
 \caption{Ablation studies on different schema representations.}
 \label{tab:ablation_schema}
    \begin{center}
    \begin{minipage}{0.8\textwidth}
    \begin{tabular}{cccc}
    \toprule
    Model & EX(DDL Schema, \%) &  EX(MAC-SQL Schema, \%) & EX(M-Schema, \%)  \\ 
    \midrule
    GPT-4o & 55.67 &57.30 & \textbf{57.95} \\ 
    DeepSeek & 53.52  & \textbf{55.28} & 55.15 \\ 
    Claude 3.5 Sonnet & 49.74 & 50.26 & \textbf{51.04} \\
    Gemini 1.5 Pro & 49.22 & \textbf{52.41} & 52.15 \\
    \bottomrule
    \end{tabular}
    \end{minipage}
    \end{center}
    
\end{table}

\subsubsection{Schema Linking}
We conduct ablation study to evaluate the effectiveness of schema linking.
We utilize recall and precision metrics to evaluate the correctness of the selected columns based on the corrected SQL query, which serves as the ground truth.
We use GPT-4o as the NL2SQL generator to analyze the impact of schema linking on end-to-end EX metrics.
The results are show in Table~\ref{tab:ablation_schema_linking}.
Without schema linking, we provide all tables, columns and random sampled example values to LLM.
It shows a precision of 10.14\% and EX of 57.95\%.
The schema linking method in this report achieves a high precision of 74.74\% while only slightly decreasing the recall.
By providing the most relevant information to the model, the execution accuracy is improved by 2.15\%, demonstrating the effectiveness of schema linking.

\begin{table}
 \caption{Ablation studies on schema linking.}
 \vspace{0.5em}
\centering\begin{minipage}{0.8\linewidth}
     \centering\begin{tabular}{cccc}
    \toprule
    Method & Precision(\%) & Recall(\%) & EX(\%) \\ \midrule
    Baseline & 10.14 & 100.00 & 57.95 \\ 
    + Schema Linking & 74.74 & 95.47 & 60.10  \\ 
    \bottomrule
    \end{tabular}
    \end{minipage}
    \label{tab:ablation_schema_linking}
\end{table}

\begin{table}[!tp]
 \caption{Ablation studies of candidate generation and selection on the performance of XiYan-SQL on the Bird development benchmark.}
    \begin{center}
    \begin{minipage}{0.5\textwidth}
    \begin{tabular}{ccc}
    \toprule
    Method &  EX(\%) & $\Delta$EX(\%) \\    
    \midrule
    XiYan-SQL All & 71.58 & -\\ 
    XiYan-SQL w/o Fine-tuned generator & 68.67 &-2.91 \\
    XiYan-SQL w/o ICL generator & 70.27  &-1.31 \\
    XiYan-SQL w/o Refiner & 71.03  &-0.55\\ 
    XiYan-SQL w/o Selection model & 68.84 &-2.74\\    
    \midrule
    XiYan-SQL All (five candidates) & 73.34 & -\\ 
    \bottomrule
    \end{tabular}
    \end{minipage}
    \end{center}
    \label{tab7}
\end{table}

\subsubsection{Candidate Generation and Selection}
To evaluate the effectiveness and impact of candidate generation and selection, we conduct various ablation studies on XiYan-SQL. 
Table \ref{tab7} presents the performance of XiYan-SQL when certain components are dropped, highlighting their significance in achieving high-quality performance.
The "XiYan-SQL All" method achieves an accuracy of 71.51\% by utilizing three candidates, of which two are generated from two distinct fine-tuned SQL generators (as described in Section \ref{section51}), while one is produced by the ICL SQL generator with GPT-4o (as presented in Section \ref{section52}).
For the candidate generator, there is a significant decrease in the performance of XiYan-SQL when the fine-tuned candidate generators are removed, further indicating that our generator is capable of generating high-quality and diverse candidate SQL queries. 
Similarly, the removal of the ICL generator and Refiner also leads to a decline in performance.
Additionally, concerning candidate selection, we observe that when the selection model is not employed, relying solely on self-consistency for candidate selection, XiYan-SQL's performance decreases by approximately three percentage points. 
This finding underscores the effectiveness of our proposed method.
Finally, when the number of SQL candidates is increased to five, the accuracy of XiYan-SQL can further reach 73.34\%.

\section{Conclusion}
In this technical report, we present a multi-generator ensemble framework for NL2SQL, named XiYan-SQL, which harnesses the benefits of the SFT approach to achieve enhanced controllability while also integrating the ICL approach to maximize the generation of high-quality and diverse SQL candidates.
We propose a two-stage and multi-task training method to train a series of models with different preferences, along with a candidate selection strategy to select the most reasonable candidate.
Xiyan-SQL demonstrates state-of-the-art performance on publicly available relational databases, including Spider and SQL-Eval, as well as on non-relational database NL2GQL.
This highlights the significant potential of XiYan-SQL for high-quality NL2SQL generation on unseen samples coming from different distributions.

\bibliographystyle{plain}
\bibliography{references}

\begin{thebibliography}{10}

\bibitem{gpt4}
Josh Achiam, Steven Adler, Sandhini Agarwal, Lama Ahmad, Ilge Akkaya, Florencia~Leoni Aleman, Diogo Almeida, Janko Altenschmidt, Sam Altman, Shyamal Anadkat, et~al.
\newblock Gpt-4 technical report.
\newblock {\em arXiv preprint arXiv:2303.08774}, 2023.

\bibitem{chen2023text}
Ziru Chen, Shijie Chen, Michael White, Raymond Mooney, Ali Payani, Jayanth Srinivasa, Yu~Su, and Huan Sun.
\newblock Text-to-sql error correction with language models of code.
\newblock {\em arXiv preprint arXiv:2305.13073}, 2023.

\bibitem{dong2023c3}
Xuemei Dong, Chao Zhang, Yuhang Ge, Yuren Mao, Yunjun Gao, Jinshu Lin, Dongfang Lou, et~al.
\newblock C3: Zero-shot text-to-sql with chatgpt.
\newblock {\em arXiv preprint arXiv:2307.07306}, 2023.

\bibitem{llama3}
Abhimanyu Dubey, Abhinav Jauhri, Abhinav Pandey, Abhishek Kadian, Ahmad Al-Dahle, Aiesha Letman, Akhil Mathur, Alan Schelten, Amy Yang, Angela Fan, et~al.
\newblock The llama 3 herd of models.
\newblock {\em arXiv preprint arXiv:2407.21783}, 2024.

\bibitem{DAIL}
Dawei Gao, Haibin Wang, Yaliang Li, Xiuyu Sun, Yichen Qian, Bolin Ding, and Jingren Zhou.
\newblock Text-to-sql empowered by large language models: A benchmark evaluation.
\newblock {\em arXiv preprint arXiv:2308.15363}, 2023.

\bibitem{gu2023few}
Zihui Gu, Ju~Fan, Nan Tang, Lei Cao, Bowen Jia, Sam Madden, and Xiaoyong Du.
\newblock Few-shot text-to-sql translation using structure and content prompt learning.
\newblock {\em Proceedings of the ACM on Management of Data}, 1(2):1--28, 2023.

\bibitem{mcssql}
Dongjun Lee, Choongwon Park, Jaehyuk Kim, and Heesoo Park.
\newblock Mcs-sql: Leveraging multiple prompts and multiple-choice selection for text-to-sql generation.
\newblock {\em arXiv preprint arXiv:2405.07467}, 2024.

\bibitem{supersql}
Boyan Li, Yuyu Luo, Chengliang Chai, Guoliang Li, and Nan Tang.
\newblock The dawn of natural language to sql: Are we fully ready?
\newblock {\em arXiv preprint arXiv:2406.01265}, 2024.

\bibitem{codes}
Haoyang Li, Jing Zhang, Hanbing Liu, Ju~Fan, Xiaokang Zhang, Jun Zhu, Renjie Wei, Hongyan Pan, Cuiping Li, and Hong Chen.
\newblock Codes: Towards building open-source language models for text-to-sql.
\newblock {\em Proceedings of the ACM on Management of Data}, 2(3):1--28, 2024.

\bibitem{bird}
Jinyang Li, Binyuan Hui, Ge~Qu, Jiaxi Yang, Binhua Li, Bowen Li, Bailin Wang, Bowen Qin, Ruiying Geng, Nan Huo, et~al.
\newblock Can llm already serve as a database interface? a big bench for large-scale database grounded text-to-sqls.
\newblock {\em Advances in Neural Information Processing Systems}, 36, 2024.

\bibitem{petsql}
Zhishuai Li, Xiang Wang, Jingjing Zhao, Sun Yang, Guoqing Du, Xiaoru Hu, Bin Zhang, Yuxiao Ye, Ziyue Li, Rui Zhao, et~al.
\newblock Pet-sql: A prompt-enhanced two-stage text-to-sql framework with cross-consistency.
\newblock {\em arXiv preprint arXiv:2403.09732}, 2024.

\bibitem{2024arXiv241018406L}
Zhisheng {Lin}, Yifu {Liu}, Zhiling {Luo}, Jinyang {Gao}, and Yu~{Li}.
\newblock {MoMQ: Mixture-of-Experts Enhances Multi-Dialect Query Generation across Relational and Non-Relational Databases}.
\newblock {\em arXiv e-prints}, page arXiv:2410.18406, October 2024.

\bibitem{liu2022semantic}
Aiwei Liu, Xuming Hu, Li~Lin, and Lijie Wen.
\newblock Semantic enhanced text-to-sql parsing via iteratively learning schema linking graph.
\newblock In {\em Proceedings of the 28th ACM SIGKDD Conference on Knowledge Discovery and Data Mining}, pages 1021--1030, 2022.

\bibitem{liu2024deepseek}
Aixin Liu, Bei Feng, Bin Wang, Bingxuan Wang, Bo~Liu, Chenggang Zhao, Chengqi Dengr, Chong Ruan, Damai Dai, Daya Guo, et~al.
\newblock Deepseek-v2: A strong, economical, and efficient mixture-of-experts language model.
\newblock {\em arXiv preprint arXiv:2405.04434}, 2024.

\bibitem{liu2024survey}
Xinyu Liu, Shuyu Shen, Boyan Li, Peixian Ma, Runzhi Jiang, Yuyu Luo, Yuxin Zhang, Ju~Fan, Guoliang Li, and Nan Tang.
\newblock A survey of nl2sql with large language models: Where are we, and where are we going?
\newblock {\em arXiv preprint arXiv:2408.05109}, 2024.

\bibitem{distillery}
Karime Maamari, Fadhil Abubaker, Daniel Jaroslawicz, and Amine Mhedhbi.
\newblock The death of schema linking? text-to-sql in the age of well-reasoned language models.
\newblock {\em arXiv preprint arXiv:2408.07702}, 2024.

\bibitem{chasesql}
Mohammadreza Pourreza, Hailong Li, Ruoxi Sun, Yeounoh Chung, Shayan Talaei, Gaurav~Tarlok Kakkar, Yu~Gan, Amin Saberi, Fatma Ozcan, and Sercan~O Arik.
\newblock Chase-sql: Multi-path reasoning and preference optimized candidate selection in text-to-sql.
\newblock {\em arXiv preprint arXiv:2410.01943}, 2024.

\bibitem{dinsql}
Mohammadreza Pourreza and Davood Rafiei.
\newblock Din-sql: Decomposed in-context learning of text-to-sql with self-correction.
\newblock {\em Advances in Neural Information Processing Systems}, 36, 2024.

\bibitem{dts_sql}
Mohammadreza Pourreza and Davood Rafiei.
\newblock Dts-sql: Decomposed text-to-sql with small large language models.
\newblock {\em arXiv preprint arXiv:2402.01117}, 2024.

\bibitem{sqlgen}
Mohammadreza Pourreza, Ruoxi Sun, Hailong Li, Lesly Miculicich, Tomas Pfister, and Sercan~O Arik.
\newblock Sql-gen: Bridging the dialect gap for text-to-sql via synthetic data and model merging.
\newblock {\em arXiv preprint arXiv:2408.12733}, 2024.

\bibitem{tasql}
Ge~Qu, Jinyang Li, Bowen Li, Bowen Qin, Nan Huo, Chenhao Ma, and Reynold Cheng.
\newblock Before generation, align it! a novel and effective strategy for mitigating hallucinations in text-to-sql generation.
\newblock {\em arXiv preprint arXiv:2405.15307}, 2024.

\bibitem{rai2023improving}
Daking Rai, Bailin Wang, Yilun Zhou, and Ziyu Yao.
\newblock Improving generalization in language model-based text-to-sql semantic parsing: Two simple semantic boundary-based techniques.
\newblock {\em arXiv preprint arXiv:2305.17378}, 2023.

\bibitem{sun2023sql}
Ruoxi Sun, Sercan~{\"O} Arik, Alex Muzio, Lesly Miculicich, Satya Gundabathula, Pengcheng Yin, Hanjun Dai, Hootan Nakhost, Rajarishi Sinha, Zifeng Wang, et~al.
\newblock Sql-palm: Improved large language model adaptation for text-to-sql (extended).
\newblock {\em arXiv preprint arXiv:2306.00739}, 2023.

\bibitem{tai2023exploring}
Chang-Yu Tai, Ziru Chen, Tianshu Zhang, Xiang Deng, and Huan Sun.
\newblock Exploring chain of thought style prompting for text-to-sql.
\newblock In {\em Proceedings of the 2023 Conference on Empirical Methods in Natural Language Processing}, pages 5376--5393, 2023.

\bibitem{chess}
Shayan Talaei, Mohammadreza Pourreza, Yu-Chen Chang, Azalia Mirhoseini, and Amin Saberi.
\newblock Chess: Contextual harnessing for efficient sql synthesis.
\newblock {\em arXiv preprint arXiv:2405.16755}, 2024.

\bibitem{gemini15}
Gemini Team, Petko Georgiev, Ving~Ian Lei, Ryan Burnell, Libin Bai, Anmol Gulati, Garrett Tanzer, Damien Vincent, Zhufeng Pan, Shibo Wang, et~al.
\newblock Gemini 1.5: Unlocking multimodal understanding across millions of tokens of context.
\newblock {\em arXiv preprint arXiv:2403.05530}, 2024.

\bibitem{rat_sql}
Bailin Wang, Richard Shin, Xiaodong Liu, Oleksandr Polozov, and Matthew Richardson.
\newblock Rat-sql: Relation-aware schema encoding and linking for text-to-sql parsers.
\newblock {\em arXiv preprint arXiv:1911.04942}, 2019.

\bibitem{macsql}
Bing Wang, Changyu Ren, Jian Yang, Xinnian Liang, Jiaqi Bai, Linzheng Chai, Zhao Yan, Qian-Wen Zhang, Di~Yin, Xing Sun, et~al.
\newblock Mac-sql: A multi-agent collaborative framework for text-to-sql.
\newblock {\em arXiv preprint arXiv:2312.11242}, 2024.

\bibitem{wang2022proton}
Lihan Wang, Bowen Qin, Binyuan Hui, Bowen Li, Min Yang, Bailin Wang, Binhua Li, Jian Sun, Fei Huang, Luo Si, et~al.
\newblock Proton: Probing schema linking information from pre-trained language models for text-to-sql parsing.
\newblock In {\em Proceedings of the 28th ACM SIGKDD Conference on Knowledge Discovery and Data Mining}, pages 1889--1898, 2022.

\bibitem{wang2023selfconsistency}
Xuezhi Wang, Jason Wei, Dale Schuurmans, Quoc~V Le, Ed~H. Chi, Sharan Narang, Aakanksha Chowdhery, and Denny Zhou.
\newblock Self-consistency improves chain of thought reasoning in language models.
\newblock In {\em The Eleventh International Conference on Learning Representations}, 2023.

\bibitem{tool_sql}
Zhongyuan Wang, Richong Zhang, Zhijie Nie, and Jaein Kim.
\newblock Tool-assisted agent on sql inspection and refinement in real-world scenarios.
\newblock {\em arXiv preprint arXiv:2408.16991}, 2024.

\bibitem{spider}
Tao Yu, Rui Zhang, Kai Yang, Michihiro Yasunaga, Dongxu Wang, Zifan Li, James Ma, Irene Li, Qingning Yao, Shanelle Roman, et~al.
\newblock Spider: A large-scale human-labeled dataset for complex and cross-domain semantic parsing and text-to-sql task.
\newblock {\em arXiv preprint arXiv:1809.08887}, 2018.

\bibitem{r3_nl2gql}
Yuhang Zhou, Yu~He, Siyu Tian, Yuchen Ni, Zhangyue Yin, Xiang Liu, Chuanjun Ji, Sen Liu, Xipeng Qiu, Guangnan Ye, et~al.
\newblock $r^3$-nl2gql: A model coordination and knowledge graph alignment approach for nl2gql.
\newblock {\em arXiv preprint arxiv:2311.01862}, 2024.

\end{thebibliography}

\newpage
\begin{appendices}
\section{SQLite Example}
In this section, we provide an example of natural language to SQLite.

\begin{figure}[!htp] 
    \centering
    \includegraphics[width=1.0\textwidth]{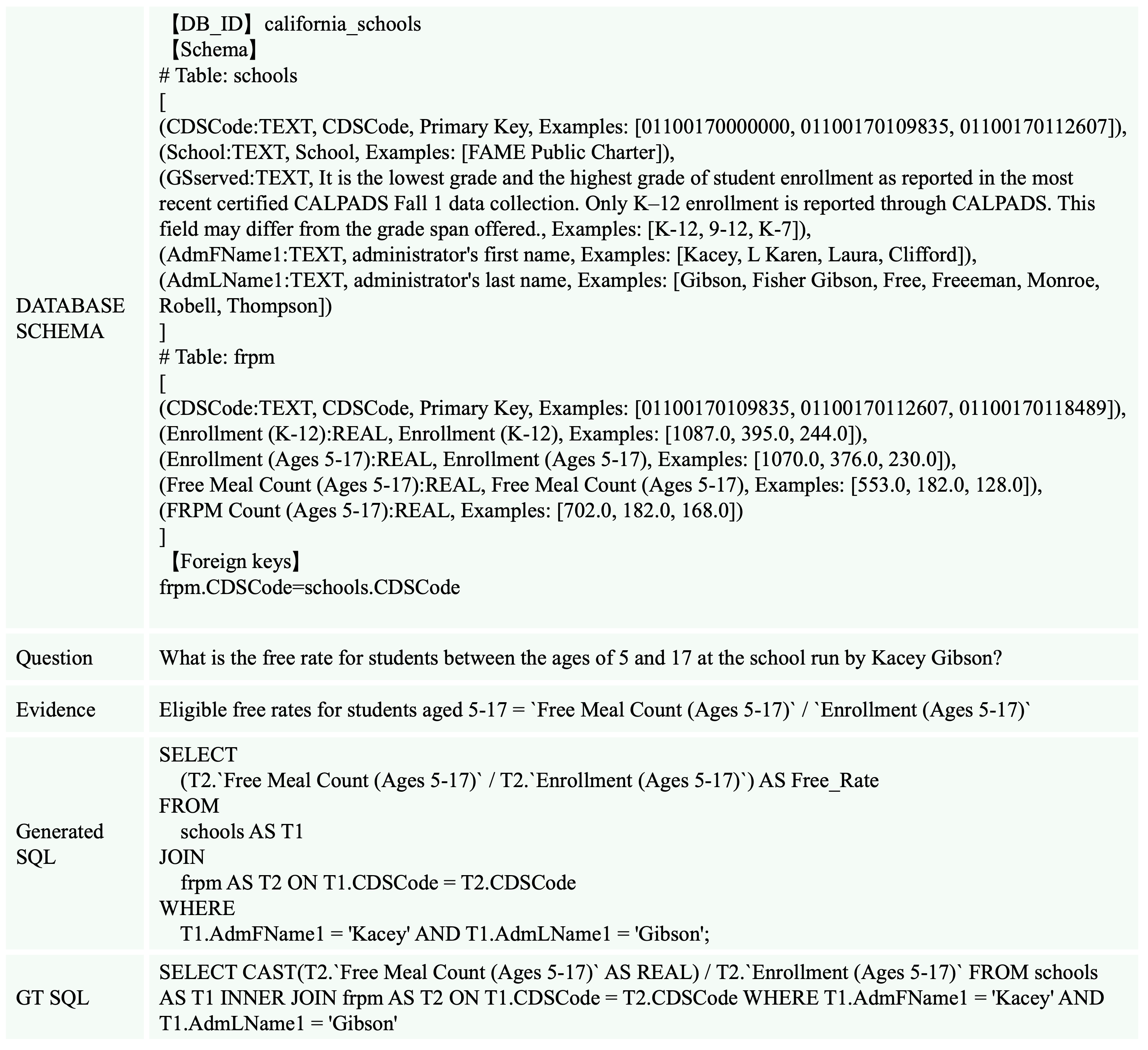}
    \caption{An example of natural language to SQLite.}
    \label{fig:nl2gql_example}
\end{figure}

\newpage
\section{PostgreSQL Example}
In this section, we provide an example of natural language to PostgreSQL.

\begin{figure}[!htp] 
    \centering
    \includegraphics[width=1.0\textwidth]{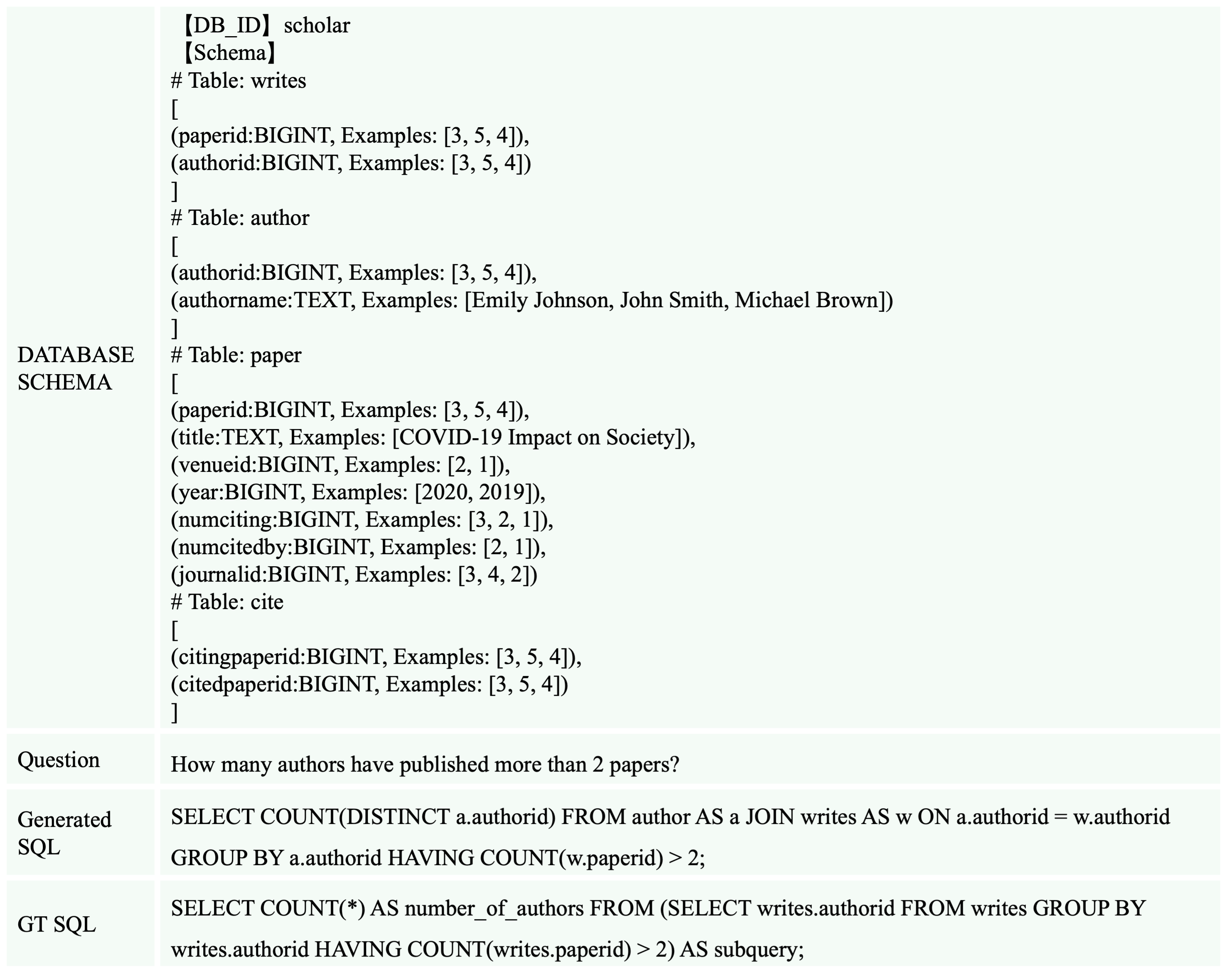}
    \caption{An example of natural language to PostgreSQL.}
    \label{fig:nl2gql_example}
\end{figure}

\newpage
\section{NL2GQL Example}
In this section, we provide an NL2GQL example. 
We extended M-Schema to represent graph databases as illustrated in Figure~\ref{fig:nl2gql_example}.

\begin{figure}[!htp] 
    \centering
    \includegraphics[width=1.0\textwidth]{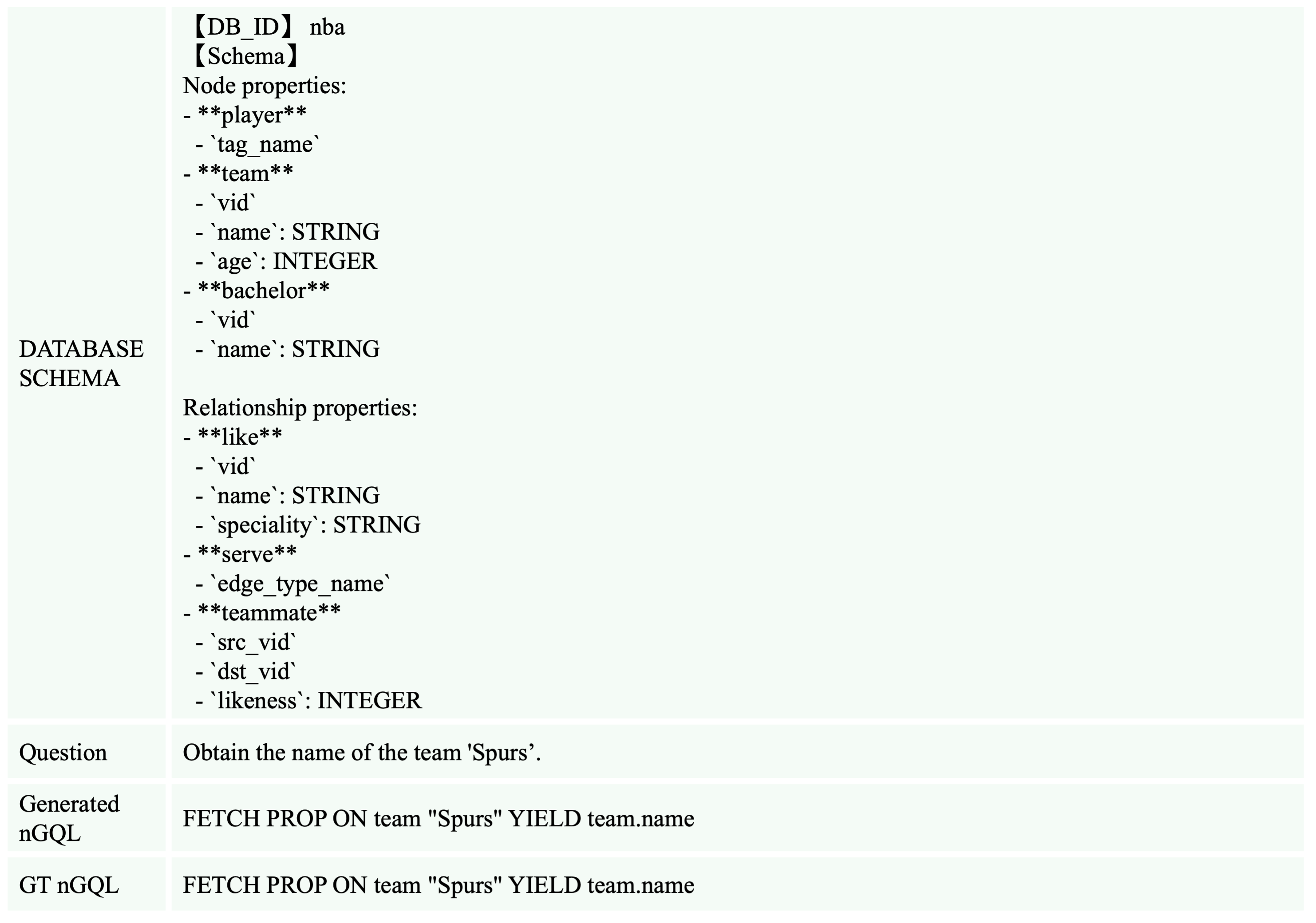}
    \caption{An example of NL2GQL.}
    \label{fig:nl2gql_example}
\end{figure}

\newpage
\section{ICL Generator Prompt}
\begin{figure}[!htp] 
    \centering
    \includegraphics[width=0.86\textwidth]{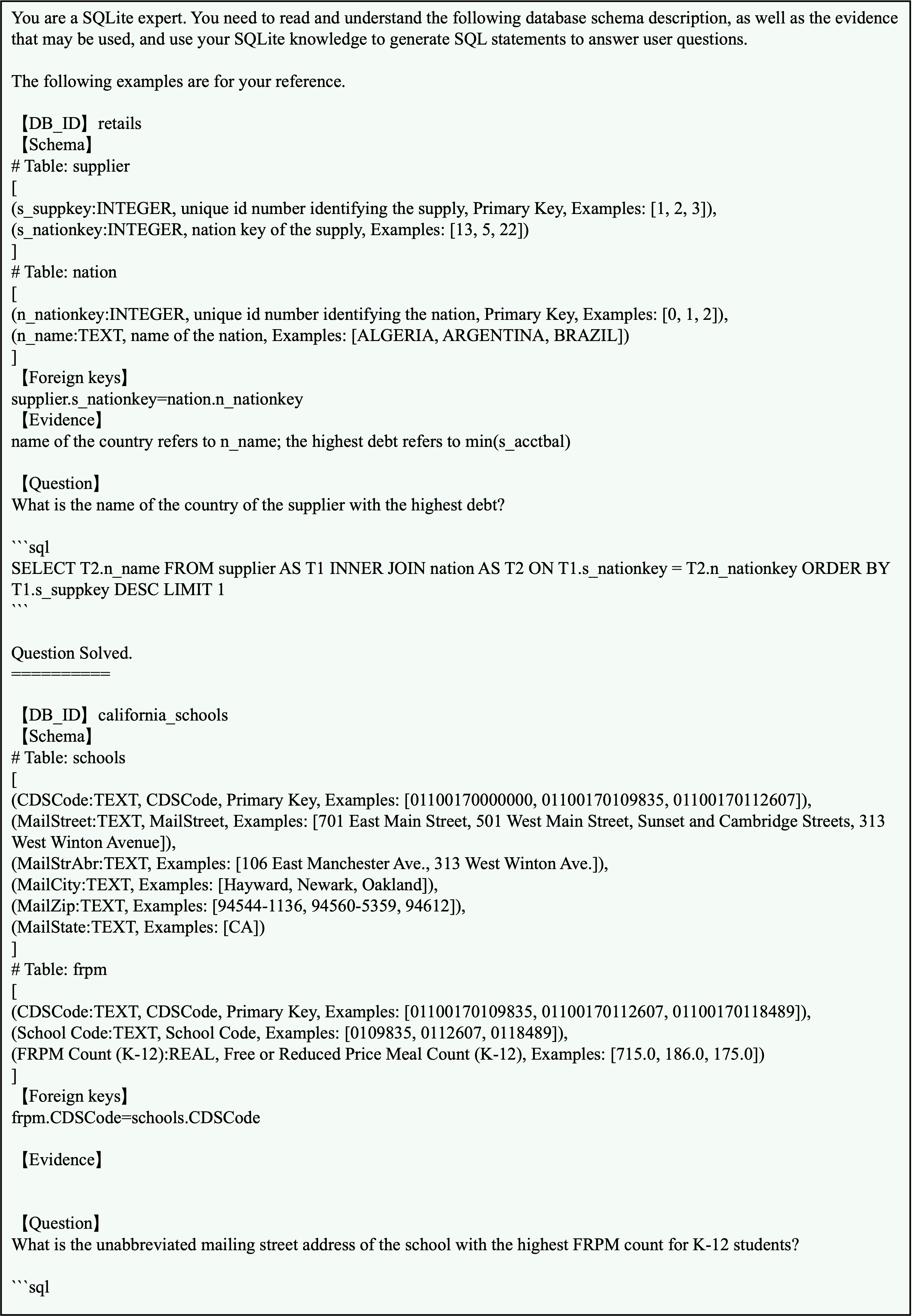}
    \caption{An example of ICL generator prompt.}
    \label{fig:ICL_prompt}
\end{figure}

In this section, we provide an example of our ICL generator prompt.
An one-shot example is presented in Figure~\ref{fig:ICL_prompt}.

\newpage
\section{Candidate Selection Prompt}
\begin{figure}[!htp] 
    \centering
    \includegraphics[width=1.0\textwidth]{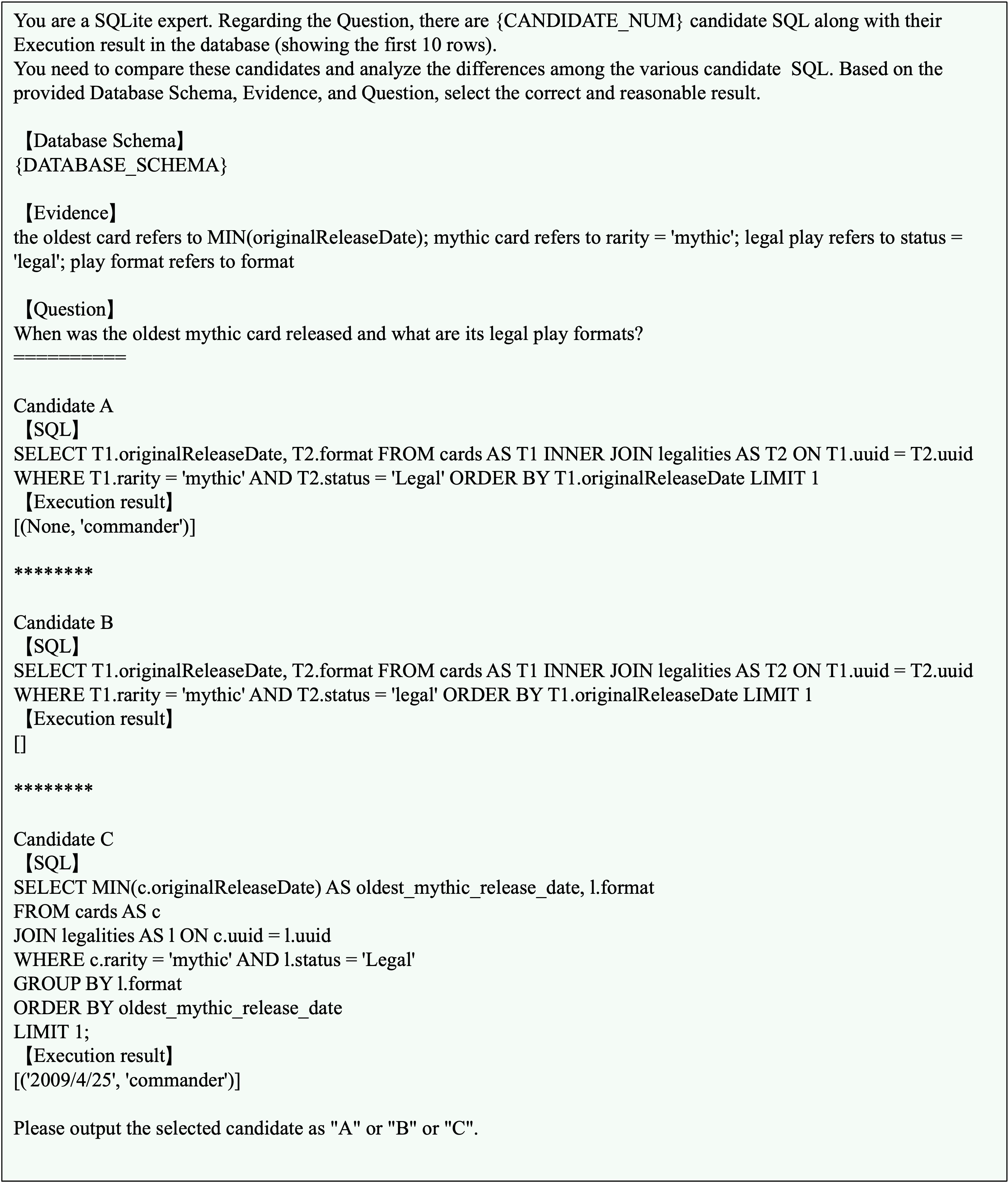}
    \caption{An example of candidate selection prompt.}
    \label{fig:candidate_selection}
\end{figure}

In this section, we provide an example of candidate selection prompt
used to pick the final response, shown in Figure~\ref{fig:candidate_selection}.

\newpage
\section{Refiner Prompt}
In this section, we provide an example prompt of Refiner used to fix syntax or logical errors.
\begin{figure}[!htp] 
    \centering
    \includegraphics[width=1.0\textwidth]{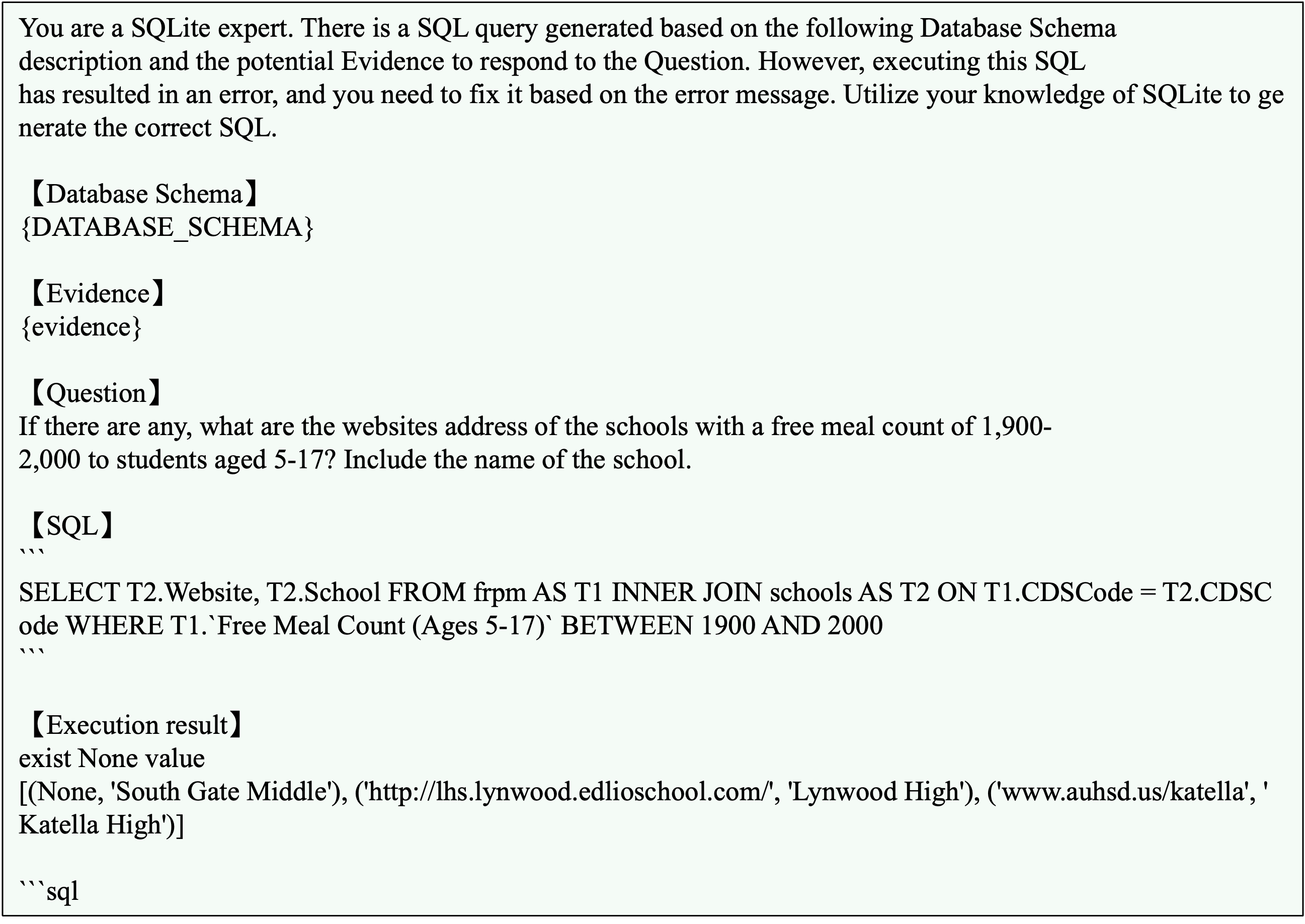}
    \caption{An example of refiner prompt.}
    \label{fig:refiner_prompt}
\end{figure}

\end{appendices}

\end{CJK}
\end{document}